\documentclass[letterpaper, 10 pt, conference]{ieeeconf}  

\IEEEoverridecommandlockouts                              

\usepackage{siunitx}
\usepackage{subcaption}
\usepackage{tikz}
\usepackage[section]{placeins}
\usepackage{fancyhdr}
\usepackage{booktabs}
\usepackage{mathtools}
\usepackage{amssymb}
\usepackage{amsmath}
\usepackage{dblfloatfix} 
\usepackage{comment}
\usepackage{hyperref}
\usepackage{graphicx}
\usepackage{xcolor}
\usepackage{nicefrac}
\usepackage{multirow}
\usepackage[ruled,vlined,linesnumbered]{algorithm2e}
\usepackage{flushend}
\usepackage{balance}
\usepackage[capitalise]{cleveref}
\usepackage[noadjust]{cite}

\def\realR{\mathbb{R}}

\def\ie{i.e.}

\DeclareMathOperator*{\argmax}{arg\,max}

\newcommand{\norm}[1]{\left\lVert#1\right\rVert}

\def\atan2{\operatorname{atan2}}

\title{\LARGE \bf
Real-Time Sampling-based Online Planning for Drone Interception
}

\author{Gilhyun Ryou$^{1}$, Lukas Lao Beyer$^{1}$ and Sertac Karaman$^{1}$%
\thanks{$^{1}$Laboratory for Information and Decision Systems, Massachusetts
Institute of Technology, Cambridge, MA. \texttt{\{ghryou, llb, sertac\}@mit.edu}}}%

\begin{document}

\maketitle
\thispagestyle{empty}
\pagestyle{empty}

\begin{abstract}
This paper studies high-speed online planning in dynamic environments. 
The problem requires finding time-optimal trajectories that conform to system dynamics, meeting computational constraints for real-time adaptation, and accounting for uncertainty from environmental changes.
To address these challenges, we propose a sampling-based online planning algorithm that leverages neural network inference to replace time-consuming nonlinear trajectory optimization, enabling rapid exploration of multiple trajectory options under uncertainty.
The proposed method is applied to the drone interception problem, where a defense drone must intercept a target while avoiding collisions and handling imperfect target predictions. 
The algorithm efficiently generates trajectories toward multiple potential target drone positions in parallel. 
It then assesses trajectory reachability by comparing traversal times with the target drone's predicted arrival time, ultimately selecting the minimum-time reachable trajectory. 
Through extensive validation in both simulated and real-world environments, we demonstrate our method's capability for high-rate online planning and its adaptability to unpredictable movements in unstructured settings.
\end{abstract}

\section{INTRODUCTION}


Rapid online trajectory adaptation is essential for deploying agile autonomous vehicles in dynamic environments. 
This capability is crucial in real-world scenarios, where unseen obstacles may appear suddenly and target positions can change unexpectedly. 
In these situations, the planning algorithm must swiftly adapt the trajectory to accommodate unforeseen obstacles or varying target points. 
This is challenging since the algorithm should efficiently handle the two critical components—obstacle avoidance and dynamic feasibility—with minimal computation to ensure quick adaptation.

\begin{figure}[tb]
    \centering
    \includegraphics[width=0.48\textwidth,trim=0.1cm 0.1cm 0.1cm 0.1cm,clip]{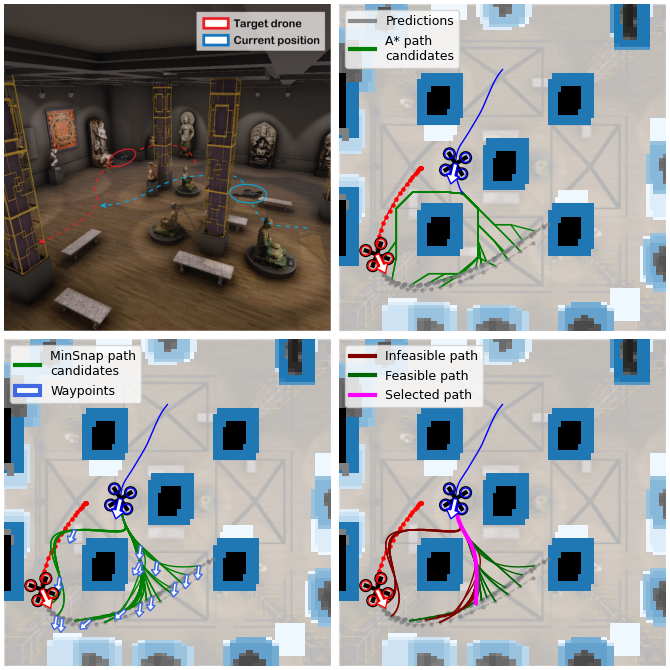}
    \vspace{-1\baselineskip}
    \caption{Overview of the proposed algorithm: 
    (top left) simulation environment for drone interception experiments;
    (top right) generation of the point-mass path towards target drone position candidates; 
    (bottom left) extraction of waypoints along these candidate paths; 
    (bottom right) parallel optimization of the trajectories along the waypoints.}
    \label{fig:alg_overview}
    \vspace{-0.5\baselineskip}
\end{figure}

Existing pathfinding methods efficiently navigate complex environments but struggle to incorporate full vehicle dynamics due to increased search space complexity. 
They often use simplified kinodynamic models, potentially resulting in infeasible trajectories during high-speed maneuvers~\cite{pivtoraiko2005efficient, liu2017search}. 
An alternative approach reduces planning complexity while considering full dynamics by using informed sampling around a point-mass initial path~\cite{penicka2022minimum}.
Nonlinear optimization methods can incorporate complex vehicle dynamics but require approximating collision avoidance constraints~\cite{liu2017planning, marcucci2024shortest, shao2023design, wang2023speed}. 
These approaches also constrain trajectory optimization to the initial path's homotopy class, which is problematic in dynamic environments requiring evaluation of multiple homotopy classes~\cite{beyer2024riskpredictive, schmittle2024multi, zhou2021raptor}. 
Furthermore, even when constrained to a single homotopy class, the final trajectory generation step remains computationally intensive and often unsuitable for high-update-rate online planning.

To address these challenges, we present a sampling-based trajectory generation method for uncertain target positions as shown in \cref{fig:alg_overview}.
Our approach builds upon recent works that replace time-consuming nonlinear optimization with neural network inference~\cite{ryou2024multi, wu2023learning, penicka2022learning, zhao2024learning}.
Utilization of these methods not only dramatically reduces computation times but also allows the exploration of different homotopy classes by optimizing multiple trajectories simultaneously.
Based on this idea, the proposed algorithm first generates multiple initial paths towards sampled candidate target positions, followed by parallel trajectory optimization using a neural network policy. 
The algorithm is applied to the drone interception problem and generates optimal trajectories towards the predicted target drone's positions.

The main contributions of this paper are as follows:



\begin{itemize}
    \item Utilization of a neural network policy to iteratively determine time-optimal and dynamically feasible trajectories for multiple initial paths, enhancing adaptability in unstructured environments.
    \item Application to drone interception, addressing collision avoidance and imperfect target predictions by assessing trajectory reachability and selecting the minimum-time feasible path.
    \item Validation through simulations and real-world experiments, demonstrating high-rate online replanning capabilities.
\end{itemize}

\section{PRELIMINARIES}

\subsection{Minimum-time planning}

Given the vehicle's current position, $\tilde p_{\text{init}}$, and state, $\mathcal{D\,} \tilde p_{\text{init}}$, (e.g., velocity and acceleration), the minimum-time planning problem involves finding the trajectory that reaches the goal position $\tilde p_{\text{goal}}$ in the least time while satisfying all constraints.
We denote $\mathcal{P}(\mathcal{D\,} \tilde p_{\text{init}})$ as the set of dynamically feasible trajectories starting from the initial vehicle state $\mathcal{D\,} \tilde p_{\text{init}}$. 
These are trajectories the underlying controller can track with limited error.
This problem often includes geometric constraints, $\mathcal{F}$, such as passing through prescribed waypoints or avoiding collisions.
Based on this formulation, we can express the general trajectory generation problem as follows:
\begin{equation}
\begin{aligned}
&\underset{p}{\text{minimize}} \; T \;\;\; \text{subject to} \\
&p(T) = \tilde p_{\text{goal}},\; p(0) = \tilde p_{\text{init}},\; \mathcal{D\,} p(0) = \mathcal{D\,} \tilde p_{\text{init}}, \\
&p \in \mathcal{P}(\mathcal{D\,} \tilde p_{\text{init}}),\; p \in \mathcal{F}
\label{eqn:planning_general}
\end{aligned}
\end{equation}
where $p$ is the trajectory, defined as the set of positions over time, and $T$ is the total time taken to execute the trajectory.
$\mathcal{D\,} p$ denotes higher-order position derivatives such as velocity and acceleration. 
$\mathcal{D\,} p(0) = \mathcal{D\,} \tilde p_{\text{init}}$ sets these derivatives' initial values, defining the system's starting state.

\subsection{Minimum-snap trajectory}

Quadrotor planning often employs snap minimization for trajectory optimization~\cite{mellinger2011minimum, richter2016polynomial}.
The differential flatness property of quadrotor dynamics allows describing the quadrotor's state using position, yaw, and their derivatives. 
This method represents the trajectory as a time-parameterized function of position and yaw. 
The coefficients of the trajectory are determined by minimizing the fourth-order position derivative (snap), yaw acceleration together with the total time as follows:

\begin{equation}
\begin{aligned}
&\underset{p}{\text{minimize}} \; \rho T + \sigma\left(p\right) \;\;\; \text{subject to} \\
&p(T) = \tilde p_{\text{goal}},\; p(0) = \tilde p_{\text{init}},\; \mathcal{D\,} p(0) = \mathcal{D\,} \tilde p_{\text{init}}, \\
&p \in \mathcal{P}(\mathcal{D\,} \tilde p_{\text{init}}),\; p \in \mathcal{F}
\label{eqn:naiveminsnap}
\end{aligned}
\end{equation}
where 
\begin{equation}\label{eqn:smoothness}
\sigma(p)=\int_{0}^{T} \mu_r \norm{\frac{d^4 p_{r}}{d^4 t}}^2 + \mu_\psi \Big(\frac{d^2 p_\psi}{d^2 t}\Big)^2 dt
\end{equation}
with $\mu_r$, $\mu_\psi$ and $\rho$ are weighting parameters.
These polynomials map time to position and yaw, \ie, $p(t) = \begin{bmatrix} {{}p_r(t)}^\intercal &  p_\psi(t) \end{bmatrix}^\intercal$ ($p_r(t) \in \mathcal{C}^4, p_\psi(t) \in \mathcal{C}^2$).
$\mathcal{C}^n$ is the differentiability class where its n-th order derivatives exist and are continuous.
To maintain trajectory differentiability, the initial vehicle state includes velocity, acceleration, jerk, and snap of position, along with yaw rate and yaw acceleration:
$\mathcal{D\,} \tilde p_{\text{init}} = \begin{bmatrix} \tilde p_{\text{init},r}^{(1)} \kern2pt \cdots \kern2pt \tilde p_{\text{init},r}^{(4)} \kern2pt \tilde p_{\text{init},\psi}^{(1)} \kern2pt \tilde p_{\text{init},\psi}^{(2)} \end{bmatrix}$.
The minimization of the fourth-order derivative acts as a regularization, resulting in smoother trajectories with gradual state changes. 


Optimization variables in trajectory generation include coefficients of polynomial pieces and time allocation for each segment $\mathbf{x} = [x_{1} \cdots x_{m}]$, summing to the total trajectory time, $T = {\textstyle\sum \nolimits}_{j=1}^{m}x_{j}$.
When the time allocation is fixed, the problem simplifies to a quadratic programming (QP) problem for determining coefficients and can be solved quickly using convex optimization solvers.
For instance, in the case of waypoint connecting constraints, the problem can be solved analytically using matrix multiplications as follows:
\begin{equation}\label{eqn:minsnap_1}
\begin{gathered}
\chi(\mathbf{x}, \tilde{\mathbf{ p}}, \mathcal{D\,} \tilde p_{\text{init}}) = \underset{p=[p_r, p_\psi]}{\text{argmin}} \;\;\; \sigma\left(p\right) \;\; \text{subject to}\\
p(T) = \tilde p_{\text{goal}},\; p(0) = \tilde p_{\text{init}},\; \mathcal{D\,} p(0) = \mathcal{D\,} \tilde p_{\text{init}},\; p \in \mathcal{F}
\end{gathered}
\end{equation}
where inclusion in $\mathcal{F}$ constrains the trajectory to connect prescribed waypoints $\tilde p^1\, \dots\, \tilde p^{m-1}$:
\begin{equation}
    \mathcal{F} = \{p\;|\;p({\textstyle\sum \nolimits}_{j=1}^{i}x_{j}) = \tilde{p}^i, \; i=1,\;\dots,\;m-1\},
\end{equation}
and $\tilde{ \mathbf{p}}$ is the set of all position constraints, including intermediate waypoints and the start and goal positions.
Based on this property, the snap minimization method applies a bi-level optimization approach: QP for coefficients and nonlinear optimization for time allocation. 
The nonlinear optimization determines the time allocation that minimizes the output of the QP:
\begin{equation}\label{eqn:minsnap_2}
    \begin{gathered}
\underset{\mathbf{x} \in \realR^m_{> 0}}{\text{minimize}} \;\;\; \sigma\left(\chi(\mathbf{x}, \tilde{\mathbf{ p}}, \mathcal{D\,} \tilde p_{\text{init}})\right) \\
\text{subject to}\;\; \chi(\mathbf{x}, \tilde{\mathbf{ p}}, \mathcal{D\,} \tilde p_{\text{init}}) \in \mathcal{P}(\mathcal{D\,} \tilde p_{\text{init}})
\end{gathered}
\end{equation}
This bi-level structure decomposes the high-dimensional optimization into two subproblems, improving numerical stability.

\subsection{Learning-based method}

\begin{figure}[]
\centering
\begin{subfigure}[b]{0.23\textwidth}
    \captionsetup{justification=centering}
    \includegraphics[width=\textwidth,trim=0.cm -0.5cm 0.cm 0.cm,clip]{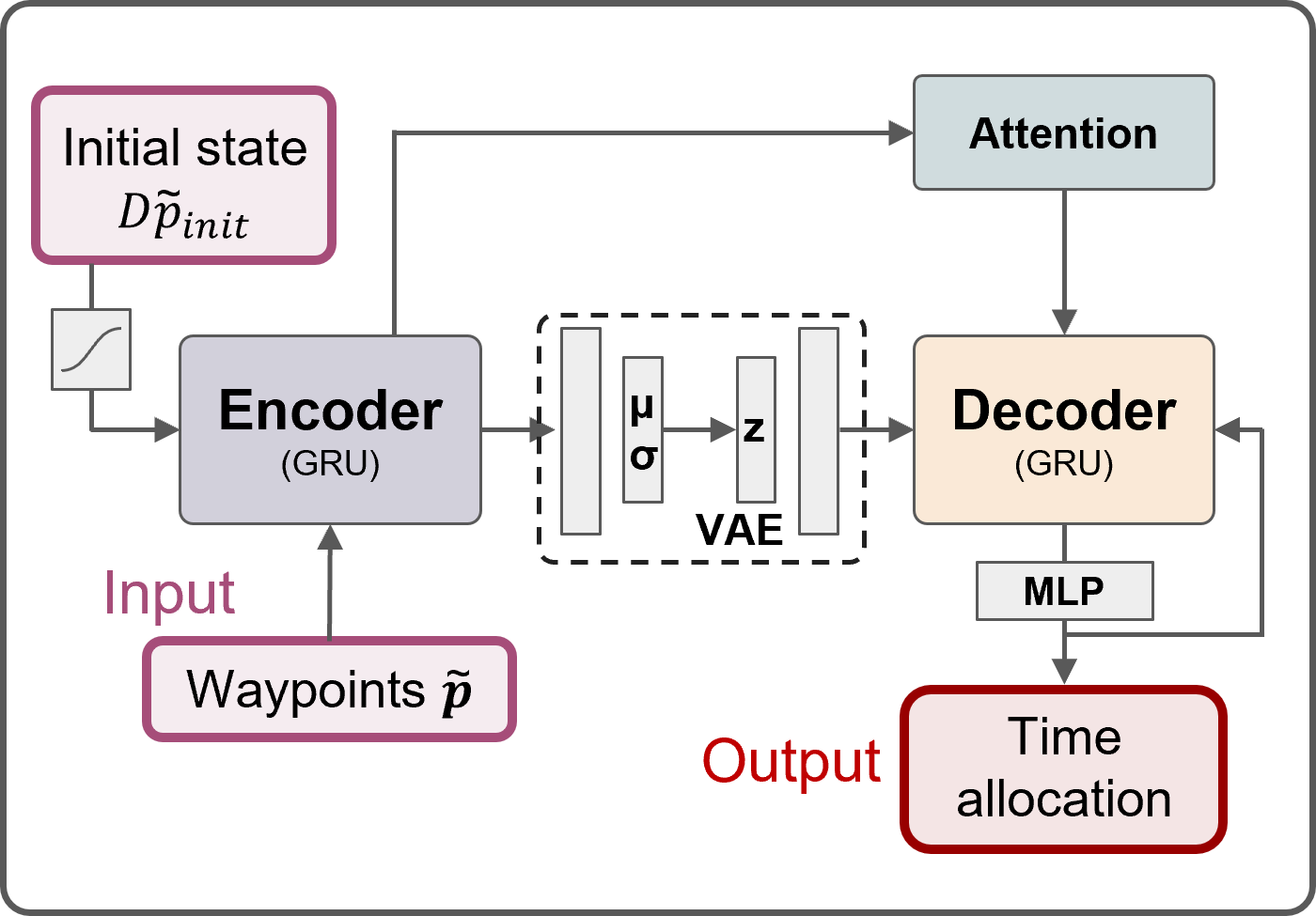}
    \vspace{-1\baselineskip}
    \caption{}
    \label{fig:policy}
\end{subfigure}
\begin{subfigure}[b]{0.23\textwidth}
    \captionsetup{justification=centering}
    \includegraphics[width=\textwidth,trim=.0cm 0.4cm .0cm .0cm,clip]{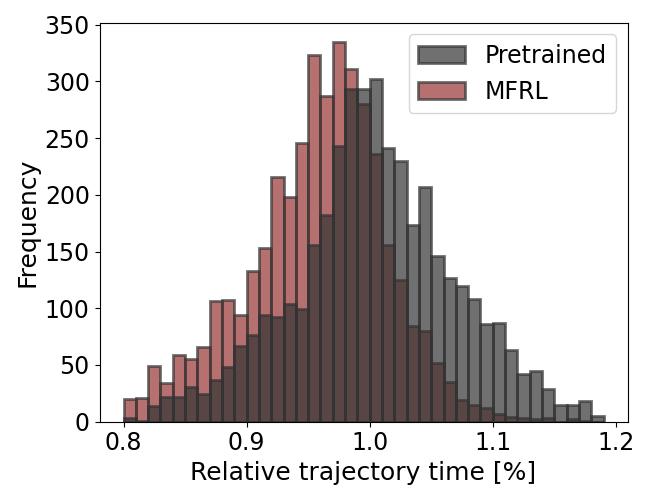}
    \vspace{-1\baselineskip}
    \caption{}
    \label{fig:policy_res}
\end{subfigure}
\caption{(a) Neural network planning policy consisting of two gated recurrent units, attention, and variational autoencoder. The model outputs time allocations from the sequence of prescribed waypoints. (b) Relative total trajectory time of the planning policy output compared to the minimum snap method. The reinforcement learning (MFRL) model outputs faster time allocations, shifting the output distribution to the left.}
\vspace{-1\baselineskip}
\end{figure}

Recent research accelerates trajectory generation by replacing costly nonlinear optimization with neural network inference. 
This approach trains a network to output optimal time allocations based on input conditions and constraints, enabling faster planning suitable for real-time scenarios in dynamic environments. 
In this work, we utilize the planning policy proposed by \cite{ryou2024multi}. 
As illustrated in \cref{fig:policy}, this method trains a sequence-to-sequence neural network model to output time allocations $\mathbf{x} = \pi (\tilde{\mathbf{p}}, \mathcal{D\,} \tilde p_{\text{init}})$ for given waypoint sequences $\tilde{\mathbf{p}}$ and initial conditions $\mathcal{D\,}\tilde{p}_{\text{init}}$.
The training process begins by generating a dataset through solving the nonlinear optimization problem \eqref{eqn:minsnap_2} for randomly sampled waypoint sequences.
The planning policy is then trained to imitate the outputs of this nonlinear optimization. 
Polynomial trajectories are subsequently obtained by solving QP in \eqref{eqn:minsnap_1} using the generated time allocations, which are then used in the tracking controller.

Building upon this pretrained model, the method further enhances the training policy using multi-fidelity reinforcement learning (MFRL). 
It incorporates a Gaussian process classifier trained to predict whether the polynomial trajectories generated from the policy output are trackable by the controller. 
The modeling efficiency is improved by augmenting the training dataset with a simple dynamics model using a multi-fidelity Gaussian process kernel. 
This feasibility prediction, $\mathbb{P} (\cdot|\mathbf{x}, \tilde{\mathbf{p}}, \mathcal{D\,} \tilde p_{\text{init}})$, serves as a reward signal, calculated as the product of relative time reduction and feasibility probability:
\begin{equation}
    r(\mathbf{x}) = \mathbb{P}(\cdot|\mathbf{x}, \tilde{\mathbf{p}}, \mathcal{D\,} \tilde p_{\text{init}})\; \delta_{\text{time}}(\mathbf{x})
\end{equation}
where the relative time reduction $\delta_{time}$ is obtained by comparing the time allocation $\mathbf{x}_{t}$ with the minimium-snap time allocation $\mathbf{x}^{\text{MS}}$: 
\begin{align}
\delta_{\text{time}}(\mathbf{x}) &= 1 - {\sum \nolimits}_{i=1}^m x_{i} / {\sum \nolimits}_{i=1}^m x^{\text{MS}}_{i}.
\end{align}
As illustrated in \cref{fig:policy_res}, the pretrained model generates time allocations similar to those of the optimal minimum snap trajectories. 
The reinforcement learning (MFRL) policy outputs faster time allocations by modeling a feasibility boundary.

\section{ALGORITHM}

The objective of this algorithm is to rapidly determine a feasible, time-optimal trajectory under uncertain goal constraints. 
Our focus is on drone interception scenarios, where a defense drone must intercept a target drone flying in the same cluttered environment. 
The position of the target drone is uncertain and must be determined using a prediction model. 
We account for the realistic scenario where predictions are imperfect and time-varying, requiring rapid adaptation of the defense drone's trajectory based on the target drone’s movements.

We assume that the distribution of goal positions is obtained from the prediction model, and candidate target positions are sampled from this distribution. 
Specifically, goal candidates $[\tilde p_{\text{goal}}^1, T_{\text{goal}}^1], \dots, [\tilde p_{\text{goal}}^K, T_{\text{goal}}^K]$ are selected considering both position and arrival time.
Based on these predictions of the target drone's position, we formulate the optimization of the defense drone's trajectory $p$ as follows:

\begin{equation}
\begin{aligned}
&\underset{p, k}{\text{minimize}} \; T_{\text{goal}}^k \;\;\; \text{subject to}\\
&p(T_{\text{goal}}^k) = \tilde p_{\text{goal}}^k, \; p(0) = \tilde p_{\text{init}},\; \mathcal{D\,} p(0) = \mathcal{D\,} \tilde p_{\text{init}}, \\
&p \in \mathcal{P}(\mathcal{D\,} \tilde p_{\text{init}}),\; p \in \mathcal{F}
\label{eqn:planning_homotopy}
\end{aligned}
\end{equation}



Our proposed algorithm consists of the following steps:
First, we use A* search to find the shortest path to each goal candidate generated by the target position prediction model. 
From these paths, we sample waypoint sequences using knot optimization. 
Next, our planning policy simultaneously determines time allocations for all sequences and assesses their reachability. 
Lastly, we select the reachable trajectory with the minimum traversal time for the defense drone.

\subsection{Waypoints candidates selection}

\begin{figure}[]
    \centering
    \includegraphics[width=0.48\textwidth,trim=0.15cm 2.8cm 0.15cm 1.5cm,clip]{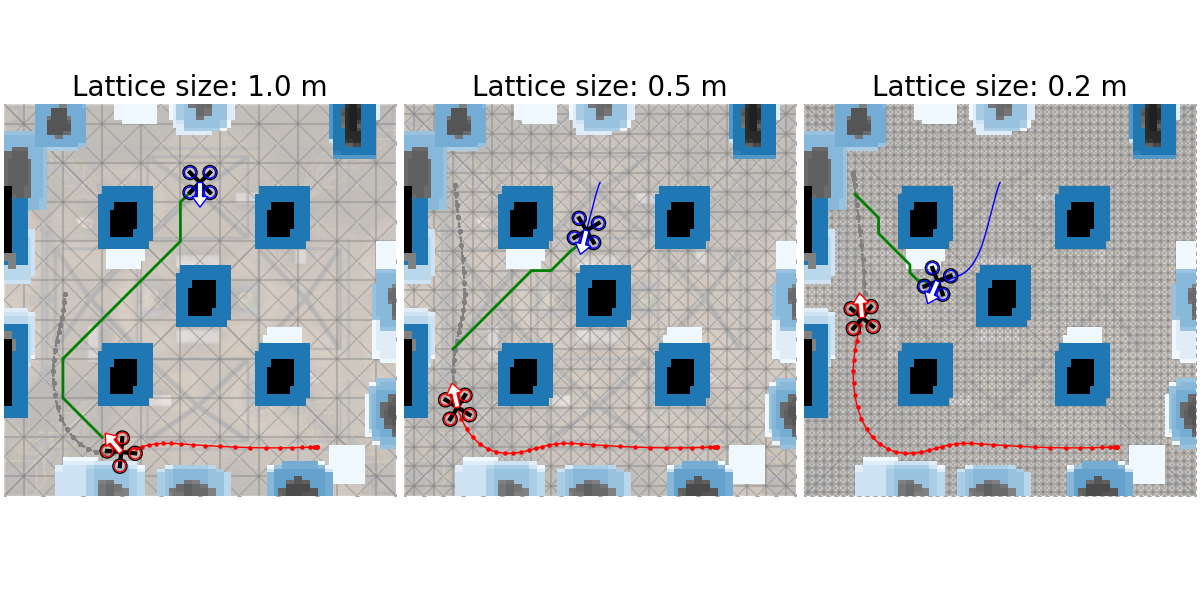}
    \caption{Utilizing a multi-resolution occupancy map for efficient A* search, where the lattice size increases with the distance between the target and defense drones.}
    \label{fig:multi_resol_astar}
    \vspace{-2.0\baselineskip}
\end{figure}

We use A* search on a multi-resolution occupancy map to determine the path to the target. 
The heuristic is directed towards the center of goal candidates' points, continuing until all candidates are reached. 
To speed up computation, we employ a dynamically adjusted grid resolution based on the distance between drones. 
The default lattice size is 0.1 m, increasing to 0.2 m, 0.5 m, and 1 m as the inter-drone distance exceeds 10 \%, 20 \%, and 40 \% of the room's diagonal, respectively, as shown in \cref{fig:multi_resol_astar}.

We select waypoints along the A* path to generate a trajectory incorporating the vehicle's dynamics model. 
These waypoints are reconnected using the minimum snap method~\cite{yeh2020fast, hernandez2003sampling}, aligning the optimized trajectory with the initial path. 
We use knot optimization, a technique from Computer-Aided Design and geometric modeling, to select appropriate endpoints for polynomial segments. Our approach combines distance and curvature parameterizations as described by \cite{pagani2018curvature}. 
To elaborate, the initial A* path $\tilde{\mathbf{q}} = [\tilde{q}^1 \cdots \tilde{q}^M]$ is reparameterized as follows:
\begin{align}
k_{d} (s) &= {\sum \nolimits}_{i=1}^{s} d(\tilde{q}^{i-1}, \tilde{q}^{i}) \\
k_{c} (s) &= {\sum \nolimits}_{i=1}^{s-1} R(\tilde{q}^{i-1}, \tilde{q}^{i}, \tilde{q}^{i+1})^{-1} (k_{d}(s+2) - k_{d}(s)) / 2 \\
k(s) &= (k_d(s) + k_c(s)) / 2
\end{align}
where $d(\cdot)$ is the Euclidean distance between adjacent points, and $R(\cdot)$ is the radius of the circle through three consecutive points.
Along the parameterization $k(s)$, we uniformly select $N_\text{wps}$ waypoints ($\tilde{\mathbf{p}} = [\tilde p^1 \cdots \tilde p^{N_\text{wps}}]$) along the A* path. 
The number of waypoints is determined as:
\begin{align}
N_{\text{wps}} = \max (k_{d}(M) / k_{d, \text{min}}, k_{c}(M) / k_{c, \text{min}})
\end{align}
where $k_{d}(M)$ and $k_{c}(M)$ represent a total distance and curvature.
$k_{d, \text{min}}$ and $k_{c, \text{min}}$ are tunable hyperparameters. 
The number of waypoints is constrained by the minimum and maximum lengths of the waypoints sequence used for training the policy, which are 2 and 14, respectively.

\subsection{Iterative traversal time adaptation}


The planning policy outputs minimum time allocations for waypoint sequences, used to generate polynomial trajectories via QP in \eqref{eqn:minsnap_1}. 
As this policy only produces minimum-time trajectories, we propose an iterative traversal time adaptation method to adjust for simultaneous arrival with the target. 
In drone interception, this approach is preferable to maximum speed, maintaining more planning options given uncertain, time-varying target positions and allowing easier direction changes.

We leverage the property that the shape of the minimum snap trajectory remains invariant when both time allocation and the current trajectory state are scaled.
To be specific, the polynomial $\chi(\alpha \mathbf{x}, \tilde {\mathbf{p}}, \alpha^{-1}\mathcal{D\,} \tilde p)$ maintains a consistent shape across all scale factors $\alpha\in\mathbb{R}_{>0}$ where $\alpha^{-1}\mathcal{D\,} \tilde p$ is defined as
\begin{equation}
\alpha^{-1} \mathcal{D\,} \tilde p \coloneqq 
\begin{bmatrix}
\alpha^{-1} \tilde{p}_r^{(1)} \kern2pt \cdots \kern2pt \alpha^{-4} \tilde{p}_r^{(4)} \kern2pt \alpha^{-1} \tilde{p}_\psi^{(1)} \kern2pt \alpha^{-2} \tilde{p}_\psi^{(2)}
\end{bmatrix}.
\label{alg:scale_property}
\end{equation}
When $\alpha > 1$, this transformation slows the trajectory while preserving its shape.

\begin{figure}[h]
\centering
\begin{subfigure}[b]{0.24\textwidth}
    \captionsetup{justification=centering}
    \begin{tikzpicture}
        \tikzset{->, arrow style/.style={draw=black, ->, >=stealth, line width=1.0pt}}
        \tikzset{->, arrow style 2/.style={draw=black, ->, >=stealth, line width=0.5pt, scale=2, dashed}}
    
        \node (A) at (0,0) {$\tilde{\mathbf{p}}, \mathcal{D\,} \tilde p$};
        \node (B) at (2.7,0) {$\mathbf{x}$};
        \node (C) at (2.7,2.3) {$\pi (\tilde{\mathbf{p}}, \alpha \mathcal{D\,} \tilde p)$};
        \node (D) at (0,2.3) {$\tilde{\mathbf{p}}, \alpha\mathcal{D\,} \tilde p$};
    
        \draw[arrow style] (C) -- (B) node[midway, right] {$\alpha^{-1}$};
        \draw[arrow style] (D) -- (C) node[midway, above] {$\pi$};
        \draw[arrow style] (A) -- (D) node[midway, left] {$\alpha$};
        \draw[arrow style 2] (A) -- (B) node[midway, above] {};
    \end{tikzpicture}
    \caption{}
    \label{fig:diagram_iter}
\end{subfigure}
\begin{subfigure}[b]{0.23\textwidth}
    \captionsetup{justification=centering}
    \includegraphics[width=\textwidth,trim=0.2cm 0.5cm .2cm .0cm,clip]{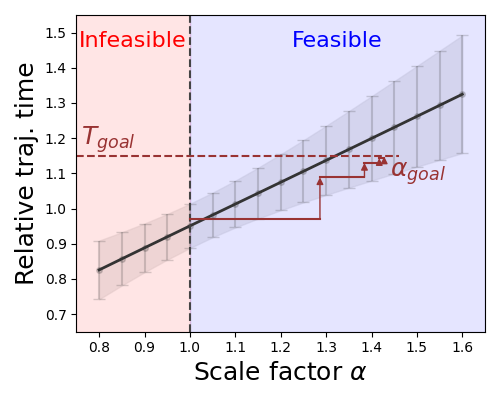}
    \caption{}
    \label{fig:plot_iter}
\end{subfigure}
\vspace{-.3\baselineskip}
\caption{Iterative traversal time adaptation: (a) Time scaling method preserves trajectory shape while reducing speed. (b) Iterative application finds optimal scaling factor $\alpha_{\text{goal}}$ to match target time.}
\vspace{-.6\baselineskip}
\end{figure}
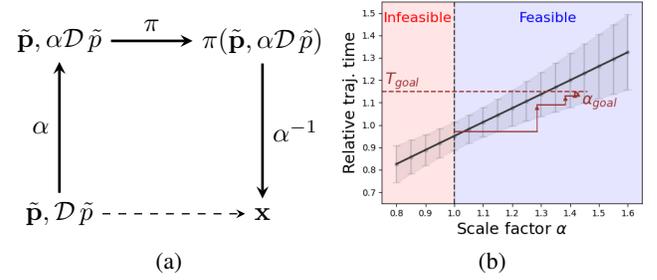

\begin{figure}[h]
    \centering
    \includegraphics[width=0.45\textwidth,trim=0.2cm 0.5cm -0.2cm 0.2cm,clip]{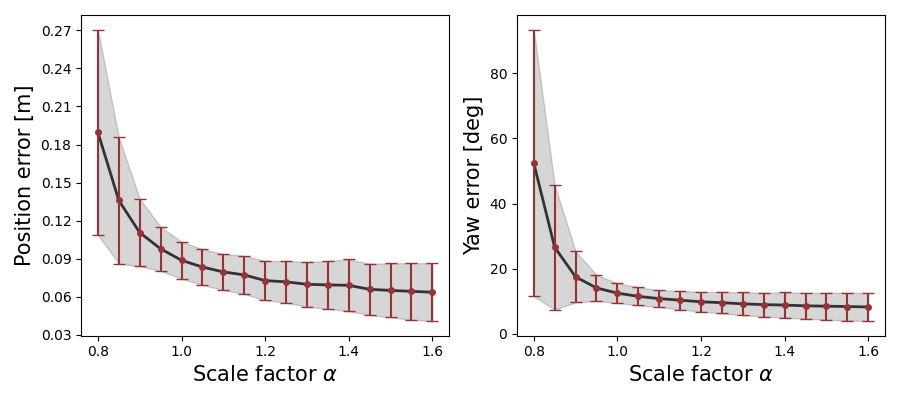}
    \caption{Tracking error depending on the scale factor. Shading indicates standard deviation.}
    \label{fig:scale_factor_err}
    \vspace{-1\baselineskip}
\end{figure}

As illustrated in \cref{fig:diagram_iter}, we propose a scaling method based on this property.
We generate scaled time allocations using the planning policy $\pi$ as follows:
\begin{equation}
    \mathbf{x} = \alpha \pi (\tilde{\mathbf{p}}, \alpha \mathcal{D\,} \tilde p)
\end{equation}
where $\alpha > 1$ speeds up the initial trajectory state $\mathcal{D\,} \tilde p$ and slows down the policy output.
This scaling method is derived by considering that the policy generates a feasible trajectory $\chi(\pi (\tilde{\mathbf{p}}, \alpha \mathcal{D\,} \tilde p), \tilde {\mathbf{p}}, \alpha \mathcal{D\,} \tilde p)$ for the initial state with $\alpha > 1$, which, when applying the scaling transformation \eqref{alg:scale_property}, yields $\chi(\alpha \pi (\tilde{\mathbf{p}}, \alpha \mathcal{D\,} \tilde p), \tilde {\mathbf{p}}, \mathcal{D\,} \tilde p)$, resulting in the same-shaped time allocations for the original initial state.
In practice, speeding up the initial state produces a more conservative policy output, which, when scaled back down, results in final time allocations that are still slower than those without scaling, reducing tracking error as shown in \cref{fig:scale_factor_err}.
Building upon this scaling method, we iteratively adjust the time allocation by slowing it down until it matches the target time $T_\text{goal}$, as shown in \cref{fig:plot_iter} and the following equation:
\begin{equation}
\alpha_\text{goal} \leftarrow \alpha_\text{goal} \times T_\text{goal} / \textstyle \sum \alpha_\text{goal} \pi (\tilde{\mathbf{p}}, \alpha_\text{goal} \mathcal{D\,} \tilde p_\text{init})
\end{equation}
The initial $\alpha_\text{goal}$ is set to $T_\text{goal} / \textstyle \sum \pi (\tilde{\mathbf{p}}, \mathcal{D\,} \tilde p_\text{init})$.
Once the scaling factor $\alpha_\text{goal}$ converges, the final time allocation is determined as $\mathbf{x} = \alpha_\text{goal} \pi (\tilde{\mathbf{p}}, \alpha_\text{goal} \mathcal{D\,}\tilde p_\text{init})$, achieving a traversing time of $T_\text{goal}$.

The planning policy is trained to generate time allocations that minimize overall traversal time. 
Therefore, by verifying that the scale factor $\alpha_\text{goal}$ is greater than one, which indicates that the target arrival time exceeds the minimal traversal time, we can assess the feasibility of each initial A* path. 
Among the optimized trajectories, we select the one with the minimal and feasible traversal time using the following equations:
\begin{equation}
\begin{gathered}
\argmax_{k} \; \tilde{\mathbb{P}}(\tilde p_{\text{goal}}^k) - T_\text{goal}^k \\ 
\text{subject to}\;\; \alpha_\text{goal}^k \geq 1, \norm{p^k- q^k}_2 < \delta_\text{max} 
\label{eqn:homotopy_1}
\end{gathered}
\end{equation}
where $\tilde{\mathbb{P}}(\tilde p_{\text{goal}}^k)$ represents the likelihood obtained from the target position prediction model, normalized to $[0,1]$ within the batch of goal position candidates.
$p^k$ is the polynomial trajectory derived from waypoints $\tilde p$ towards the goal position $\tilde p_{\text{goal}}^k$ and time allocations determined by the planning policy.
$q^k$ is obtained by linearly interpolating the A* path. 
The L2 distance between $p^k$ and $q^k$ is calculated as the maximum distance between corresponding points sampled uniformly along both paths. 
By rejecting trajectories that deviate excessively from the A* path, we ensure collision avoidance for the final polynomial trajectory.
If no feasible trajectory is available, we select the one with the longest target time, allowing us to wait until a feasible path becomes available.


\subsection{Target position prediction}
\begin{figure}[]
    \centering
    \includegraphics[width=0.48\textwidth,trim=0.15cm 2.8cm 0.15cm 1.5cm,clip]{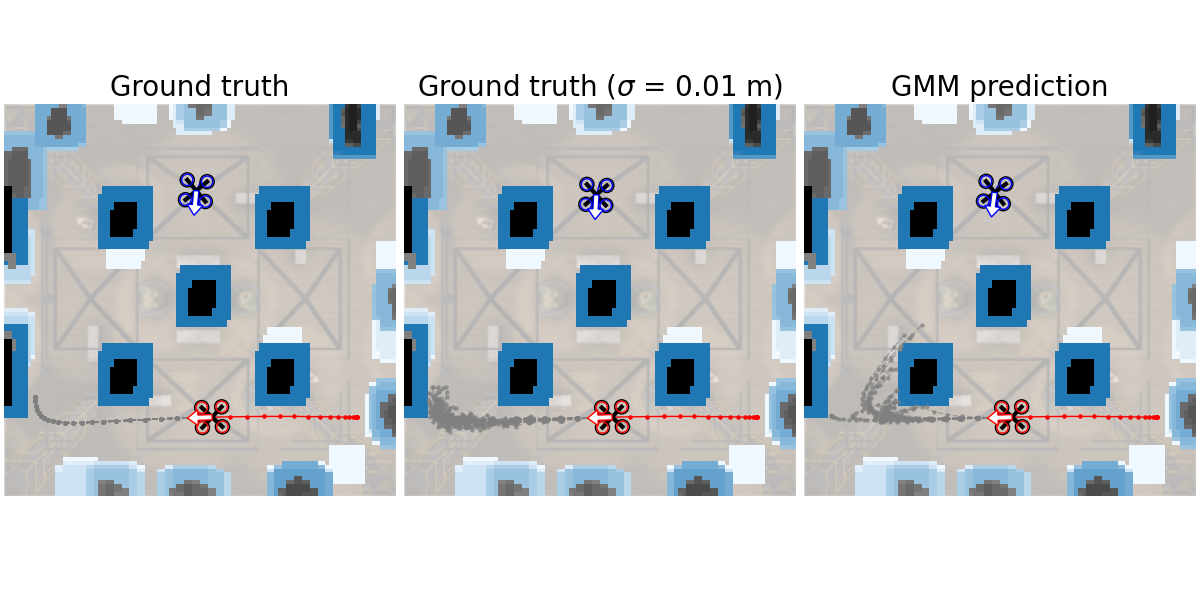}
    \caption{Comparison of prediction models.}
    \label{fig:compare_pred}
    \vspace{-1\baselineskip}
\end{figure}

We assess our algorithm using three distinct models for target position: 1) the target drone’s actual future trajectory (ground truth), 2) a noisy version of the ground truth, and 3) predictions based on a Gaussian mixture model (GMM).
The prediction model estimates the distribution of goal positions for target time $T_\text{goal}$, $\mathbb{P}(\tilde{p}_\text{goal})$.
\cref{fig:compare_pred} illustrates the three different prediction models used for evaluation.

Initially, we evaluate our algorithm with the exact ground truth trajectory. 
We uniformly select a candidate target position at each $dt$ interval for $N_\text{pred}$ steps, starting from the current time step.
This results in target position candidates $\tilde{\mathbf{p}}_{\text{target}} =[\tilde p_{\text{target}, 1}, \dots, \tilde p_{\text{target}, N_\text{pred}}]$ which corresponds to the time steps, $T_\text{goal} \in [dt, 2dt, \dots, N_\text{pred}dt]$.

Next, we introduce Gaussian noise to simulate a noisy prediction environment. 
Along the $N_\text{pred}$ steps, we $N_\text{batch}$ positions around the ground truth for each step, resulting in a total of $N_\text{pred} \times N_\text{batch}$ position candidates.
The variance of this noise increases linearly with each prediction time step to reflect growing uncertainty as follows:
\begin{equation}
    \mathbb{P}(\tilde{\mathbf{p}}_{\text{goal}}) = \mathcal{N}(\tilde{\mathbf{p}}_{\text{goal}} | \tilde{\mathbf{p}}_{\text{target}}, \text{diag}([1 \cdots N_\text{pred}]) \sigma_{pred}^2)
\end{equation}

Lastly, to implement the full planning pipeline, we train a Gaussian mixture model to predict the target drone’s trajectory distribution, which is often used to predict the trajectory of vehicles or pedestrians~\cite{wiest2012probabilistic, wakulicz2023topological, salzmann2020trajectron}.
In the planning environment, we generate random minimum snap trajectories and subsample trace points along these paths. 
We use the sequences of non-colliding points to train the prediction model. 
The distribution of sampled reference positions $\tilde{\mathbf{p}} \in \mathbb{R}^{3 \times (N_\text{obs}+N_\text{pred})}$ is modeled with a mixture of Gaussian distributions:
\begin{equation}
    \mathbb{P}(\tilde{\mathbf{p}}) = \sum_{i=1}^{N_\text{GMM}} \pi_i \mathcal{N}(\tilde{\mathbf{p}}| \mathbf\mu_i, \mathbf\Sigma_i)
\end{equation}
The Gaussian mixture model is trained using the expectation maximization (EM) algorithm, which optimizes the weights and parameters of each Gaussian distribution. 
The future position distribution $\tilde{\mathbf{p}}_{\text{goal}}$ is obtained by marginalizing observed reference positions $\mathbf{p}_{\text{obs}} \in \mathbb{R}^{3 \times N_\text{obs}}$ from the reference position distribution $\tilde{\mathbf{p}}$, based on observed target drone positions.
Similarly to the noisy ground truth scenario, $N_\text{batch}$ positions are sampled at each of the $N_\text{pred}$ time steps, and any target positions that are unreachable or located on obstacles are rejected.


\section{EXPERIMENTS}

\begin{figure*}[]
    \centering
    \includegraphics[width=0.80\textwidth,trim=.0cm 0.cm .0cm .0cm,clip]{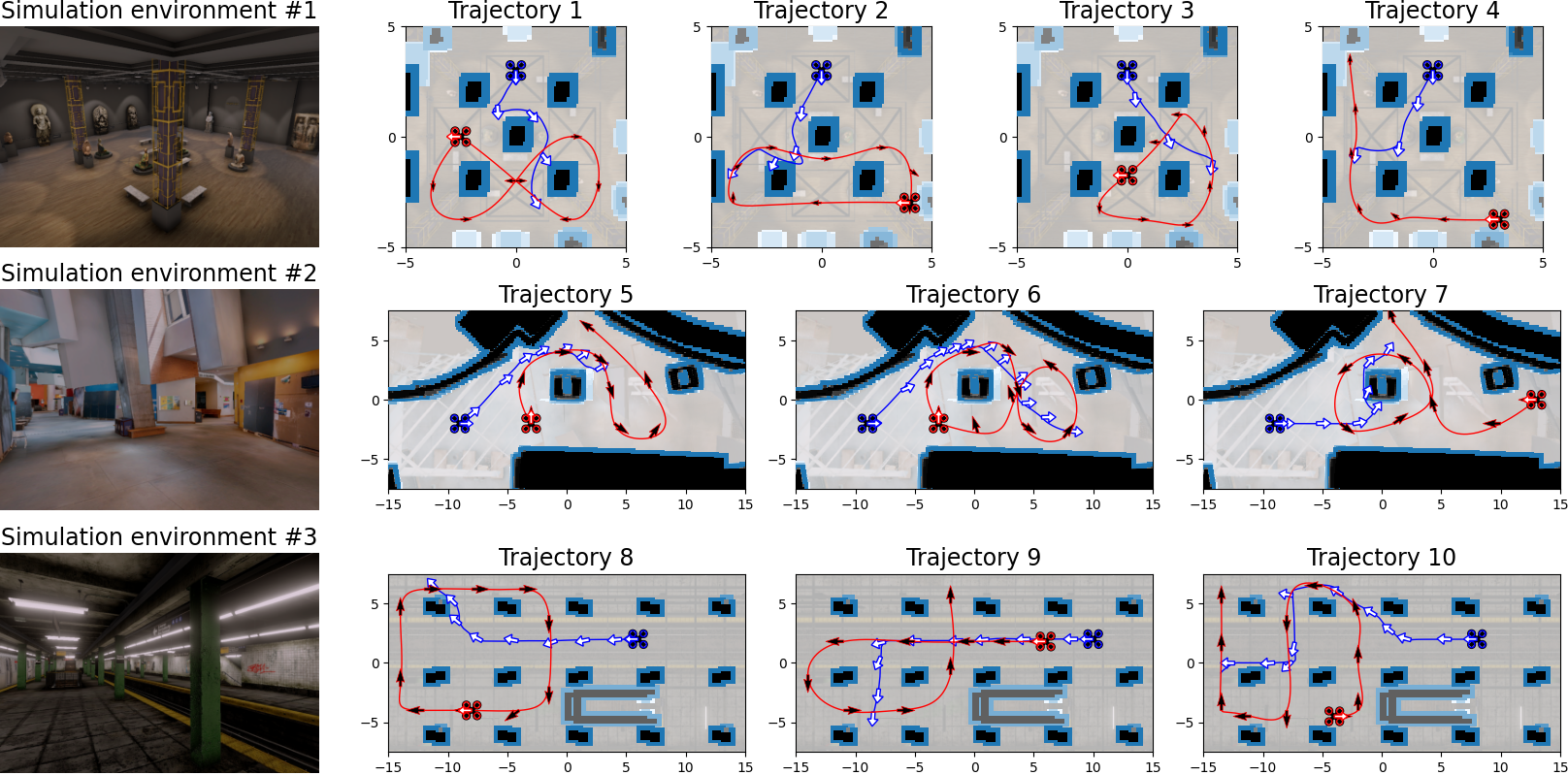}
    \caption{Simulation environments and trajectories used for evaluation. 
    Red paths represent the target drone's trajectories, and blue paths represent the defense drone's.
    }
    \label{fig:exp_env}
    \vspace{-1\baselineskip}
\end{figure*}

We evaluate the algorithm in three unique simulated environments, each with 10 different target trajectories, as shown in \cref{fig:exp_env}. 
Environment 1 has dimensions of $10 \si{m} \times 10 \si{m} \times 2 \si{m}$, while environments 2 and 3 each have dimensions of $30 \si{m} \times 15 \si{m} \times 3 \si{m}$. 
The time interval for predictions is set at $dt = 0.1 \si{s}$ for environment 1 and $dt = 0.2 \si{s}$ for environments 2 and 3, reflecting the larger spaces.
The target drone's trajectories are generated using the minimum snap method. 
For each trajectory, we vary the maximum speed of the target drone and the departure time of the defense drone. 
The 10 different maximum speeds range from $0.5 \si{m/s}$ to $5.5 \si{m/s}$, increasing uniformly by $0.5 \si{m/s}$. 
There are also 10 different departure times ranging from $0.5 \si{s}$ to $2.5 \si{s}$. 
Overall, each trajectory is evaluated using 100 different combinations of maximum speed and departure time. 
We add padding of $0.5 \si{m}$ around the obstacles in the occupancy grid, and during the selection of the optimal trajectory in \eqref{eqn:homotopy_1}, we reject trajectories that deviate more than $\delta_\text{max} = 0.5 \si{m}$ from the initial A* path.
All experiments are conducted using the FlightGoggles simulator~\cite{guerra2019flightgoggles} with the INDI trajectory tracking controller~\cite{tal2018accurate}.


We evaluate the algorithm using five prediction models: no prediction, ground truth, noisy ground truth ($\sigma_\text{pred} = 0.01 \si{m}$ and $\sigma_\text{pred} = 0.05 \si{m}$), and the GMM model. 
Without prediction, the defense drone chases the target's current position, resulting in low success rates. 
Ground truth and noisy ground truth serve to evaluate the algorithm's resilience to different levels of prediction accuracy.
The GMM model's performance falls between the noisy ground truth models with $\sigma_\text{pred} = 0.01 \si{m}$ and $\sigma_\text{pred} = 0.05 \si{m}$. 
For noisy ground truth and GMM, we sample $N_\text{batch} = 20$ positions at each $N_\text{pred} = 20$ time steps. 
GMM prediction uses $N_\text{obs} = 10$ observed target positions. 
Success is defined as the defense drone approaching within $0.4 \si{m}$ of the target, maintaining position error under $40 \si{cm}$ and yaw tracking error below 45 degrees.

As shown in \cref{tab:exp_alg_res_sim}, we compare three different planning methods over these prediction models.
The \textit{No Policy} serves as our baseline method, determining time allocation based on the distance between waypoints divided by a maximum velocity of 2.5 m/s, which was optimized for success rate.
We also examine a planning policy trained with a minimum snap trajectory (\textit{Pretrained}) and another further refined with reinforcement learning (\textit{MFRL}). 
The main limitation of the \textit{No policy} approach is its inability to generate feasible trajectories. 
By determining time allocations using the trained planning policies, this limitation can be overcome. 
Another common issue occurs when the target drone is too fast for the defense drone to intercept. 
As shown in \cref{fig:policy_res}, the use of reinforcement learning enhances the planning policy's performance, enabling faster maneuvers even with the same tracking error, thus the MFRL policy achieves a higher success rate.
\cref{tab:exp_comp_time} presents the computation times for each procedure. 
The entire process is maintained below 100 ms to ensure suitability for rapid online planning.


\begin{figure}[]
    \centering
    \includegraphics[width=.46\textwidth,trim=0.15cm 0.15cm 0.15cm .15cm,clip]{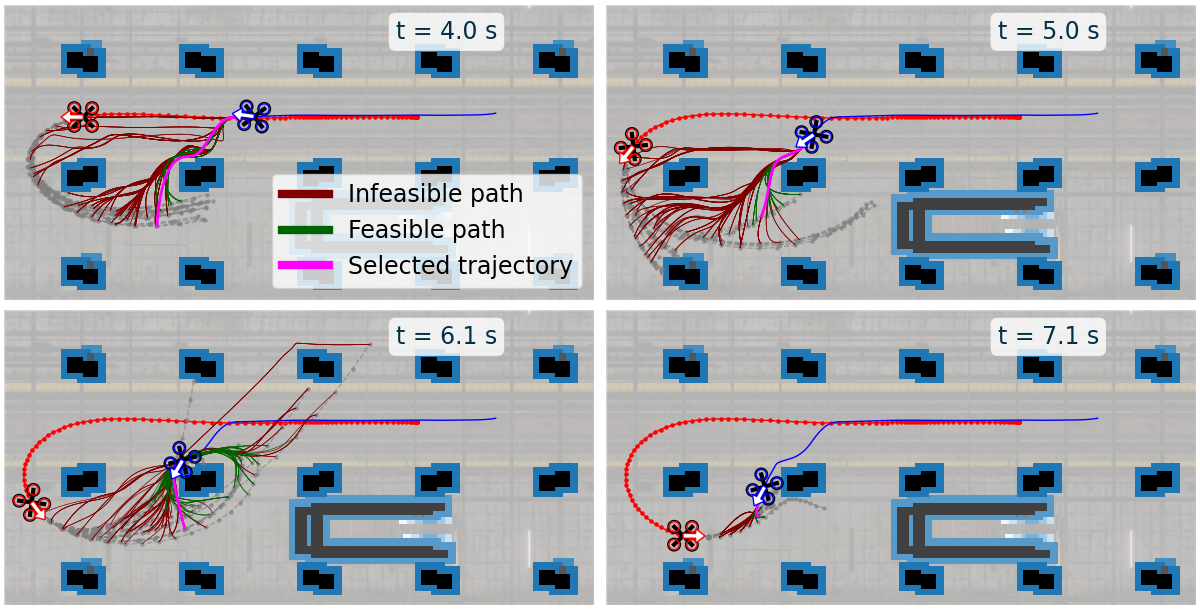}
    \caption{Optimized defense drone trajectories for each timestep. Trajectory 9 in simulation environment \#3.}
    \label{fig:exp_case_2}
\end{figure}


\begin{table}[]
\scriptsize
\centering
\caption{
    Comparison of success rate between the case without policy, pretrained policy and MFRL policy.
}
\label{tab:exp_alg_res_sim}
\begin{tabular}{c ccccc}
\toprule
\multicolumn{1}{c}{Planner} &
\multicolumn{1}{c}{\shortstack[c]{No\\prediction}} &
\multicolumn{1}{c}{\shortstack[c]{Ground\\truth}}  &
\multicolumn{1}{c}{\shortstack[c]{Ground\\truth\\($\sigma\texttt{=}0.01 \si[]{m}$)}}  &
\multicolumn{1}{c}{\shortstack[c]{Ground\\truth\\($\sigma\texttt{=}0.05 \si[]{m}$)}}  &
\multicolumn{1}{c}{\shortstack[c]{GMM\\prediction}} \\ 
\midrule
No policy      & 0.14 & \textbf{0.40} & 0.40 & 0.30 & \textbf{0.36} \\
Pretrained     & 0.38 & \textbf{0.78} & 0.76 & 0.66 & \textbf{0.68} \\
MFRL           & 0.57 & \textbf{0.90} & 0.85 & 0.76 & \textbf{0.82} \\
\hline
\end{tabular}
\vspace{-1\baselineskip}
\end{table}


\begin{table}[]
\scriptsize
\centering
\caption{
    Computation times for each procedure in the planning algorithm.
}
\label{tab:exp_comp_time}
\begin{tabular}{ccc}
\toprule
\multicolumn{1}{c}{\shortstack[c]{GMM prediction}} &
\multicolumn{1}{c}{\shortstack[c]{A* Initial path}}  &
\multicolumn{1}{c}{\shortstack[c]{Policy inference}} \\ 
\midrule
18.7 \si{ms} & 9.5 \si{ms} & 71.2 \si{ms} \\
\hline
\end{tabular}
\vspace{-1\baselineskip}
\end{table}




\cref{fig:exp_case_2} illustrates the optimized trajectories over time, where the defense drone identifies a feasible trajectory that avoids turning around the left bottom pillar and successfully reaches the target drone.
The target drone's positions are predicted using the GMM.
The algorithm dynamically updates candidate goal positions based on the target drone's movements, concurrently optimizing trajectories towards these positions and selecting the optimal trajectory.

\begin{figure}[]
    \centering
    \includegraphics[width=0.4\textwidth,trim=0.15cm 0.15cm 0.15cm 0.15cm,clip]{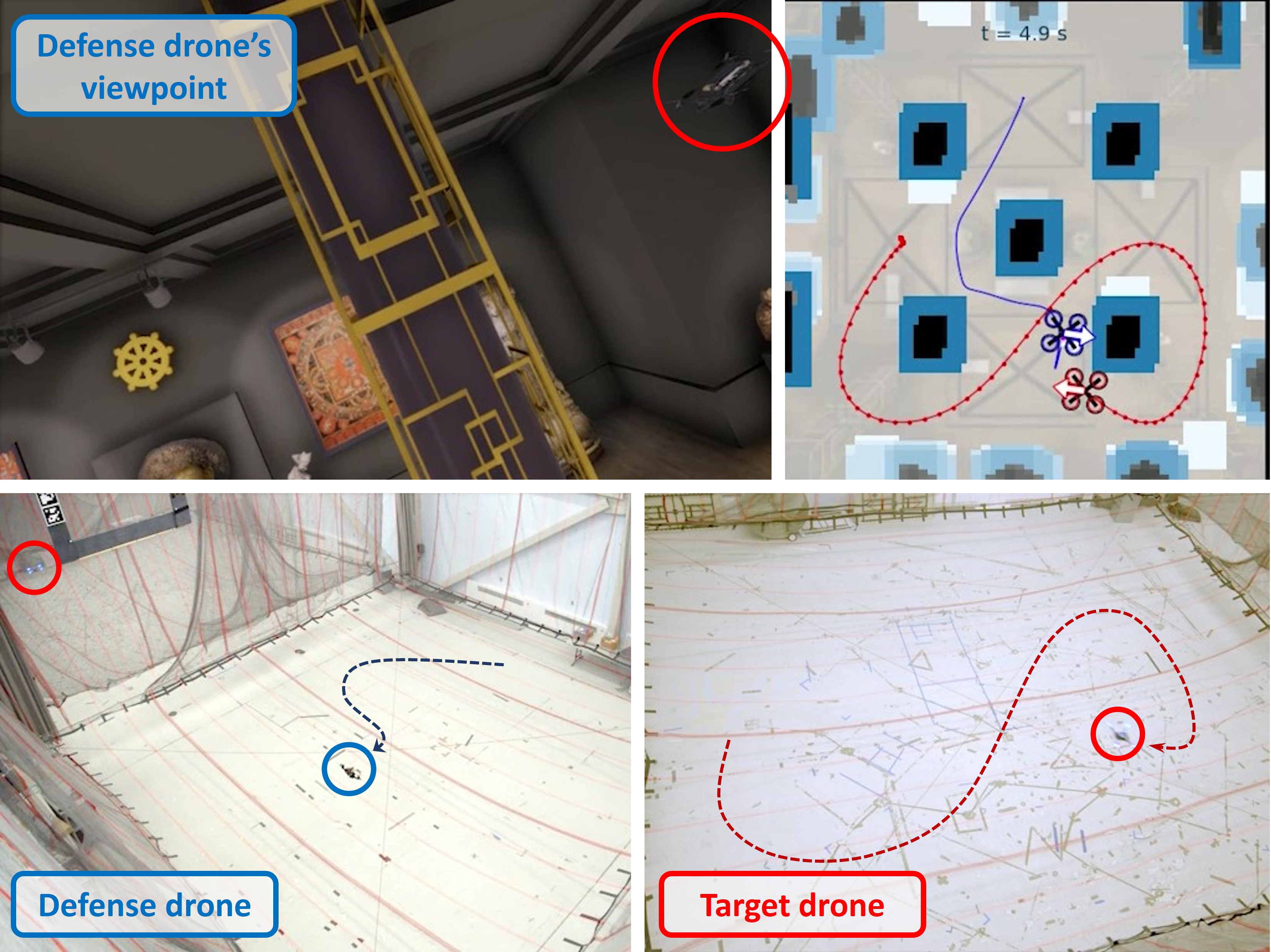}
    \caption{Real-world flight experiments demonstrating real-time sampling-based online planning for drone interception. 
    Trajectory 1 from simulation environment \#1 is used for the target drone's path. 
    The defense drone intercepts the target drone after 4.9 seconds.}
    \label{fig:exp_real}
    \vspace{-1.3\baselineskip}
\end{figure}

We further evaluate the system in real-world environments shown in \cref{fig:exp_real} using an augmented reality setup.
The target and defense drones operate in separate rooms, with their motion-captured positions rendered in a shared simulated environment to test interception scenarios safely.
The target drone's positions are predicted using GMM based on motion capture data. 
The planning policy is executed on the Titan Xp GPU in the host computer and communicates with the vehicle's microcontroller to update the trajectory.
We evaluate Trajectories 1 and 2 from the simulated environment 1, in real-world flight experiments. 
In both cases, the planner updates the trajectory at around 10 Hz and successfully intercepts the target drone.
A video of the experiments can be found at~\url{https://youtu.be/dDdshfEAZpg}.

\section{CONCLUSION}
This paper introduces a sampling-based online planning method tailored for drone interception scenarios. 
Our approach combines efficient sampling techniques with parallel trajectory optimization, leveraging a neural network policy to enhance computational speed. 
The method begins by generating multiple initial trajectories towards sampled candidate target positions. 
We employ a neural network-based planning policy for rapid trajectory optimization together with iterative traversal time adaptation, enabling high-rate online replanning towards dynamic target drones.
This method has been validated in both simulated and real-world settings, demonstrating its capability for high-rate online replanning in dynamic environments.
\newpage
\addtolength{\textheight}{-12cm}   








\bibliographystyle{IEEEtran}
\bibliography{refs}

\end{document}